\RequirePackage{fix-cm}
\documentclass{svjour3}                     
\smartqed  
\usepackage{graphicx}

\usepackage{amsmath}
\usepackage{tabularx}
\usepackage{graphicx}
\graphicspath{{images/}}

\usepackage{subfig}
\usepackage{caption}

\usepackage{multirow}
\usepackage{diagbox}
\begin{document}

\title{Multi-view Reconstructive Preserving Embedding for Dimension Reduction
}


\author{Huibing Wang \and Lin Feng \and Adong Kong \and Bo Jin 
}


\institute{Huibing Wang \at
              Information Science and Technology College, Dalian Maritime University, Dalian, China, 116024. \\
              \email{whb08421005@mail.dlut.edu.cn}           
           \and
           Lin Feng \at
           School of Computer Science and Technology, Faculty of Electronic Information and Electrical Engineering, Dalian University of Technology, Dalian, China, 116024.
           \and
           Adong Kong \at
           School of Computer Science and Technology, Faculty of Electronic Information and Electrical Engineering, Dalian University of Technology, Dalian, China, 116024.
           \and
           Bo Jin \at
           School of Computer Science and Technology, Faculty of Electronic Information and Electrical Engineering, Dalian University of Technology, Dalian, China, 116024.
}

\date{Received: date / Accepted: date}

\maketitle

\begin{abstract}
With the development of feature extraction technique, one sample always can be represented by multiple features which locate in high-dimensional space.  Multiple features can reflect various perspectives of one same sample, so there must be compatible and complementary information among the multiple views. Therefore, it's natural to integrate multiple features together to obtain better performance. However, most multi-view dimension reduction methods cannot handle multiple features from nonlinear space with high dimensions. To address this problem, we propose a novel multi-view dimension reduction method named Multi-view Reconstructive Preserving Embedding (MRPE) in this paper. MRPE reconstructs each sample by utilizing its k nearest neighbors. The similarities between each sample and its neighbors are primely mapped into lower-dimensional space in order to preserve the underlying neighborhood structure of the original manifold. MRPE fully exploits correlations between each sample and its neighbors from multiple views by linear reconstruction. Furthermore, MRPE constructs an optimization problem and derives an iterative procedure to obtain the low-dimensional embedding. Various evaluations based on the applications of document classification, face recognition and image retrieval demonstrate the effectiveness of our proposed approach on multi-view dimension reduction.
\end{abstract}




\section{Introduction}

Nowadays, along with the development of information technology, a large plenty of tools are produced to describe various samples from different perspectives. Therefore, one same sample can always be represented by multiple views which consist of compatible and complementary information. Take computer vision for example, one image can be represented by multiple feature descriptors which describes images from different perspectives, such as Scale-invariant feature transform (SIFT)\cite{lowe1999object}, Histogram of oriented gradients (HOG)\cite{dalal2005histograms} and Speeded up robust features (SURF)\cite{bay2008speeded}, etc\cite{Yu2018Multi}. Observing that these feature descriptors are capable of describing multiple properties of one image, they must have some interior relations between each other. Therefore, it is natural for researchers to integrate them together and obtain better performance rather than rely on just one single view\cite{liu2013multi}. We consider effectively exploring and exploiting multiple representations simultaneously, so the key of multi-view learning is to leverage the complementary information from multiple views, which is of vital importance but challenging.

Since multi-view learning has attracted more and more attentions, a great deal of efforts \cite{Wang2018Beyond,Wang2016Iterative,Wang2017Unsupervised} have been carried out on multi-view data learning, including classification\cite{lian2006multi,wang2017multi,feng2017spectral}, clustering\cite{kumar2011co}, and feature selection\cite{xie2016multi,wang2016multi}. Xia et al.\cite{xia2010multiview} proposed a multi-view spectral embedding (MSE) which can better understand correlations between each two samples in original space. MSE is capable of insuring that the distribution of each view is sufficiently smooth, which improves the performance of relative applications based on clustering. Xu et al.\cite{xu2015multi} adopted a novel probabilistic smoothed weighting scheme to construct a novel algorithm named Multi-view Self-Packed Learning (MSPL), which can also perform excellently in many fields (including human motion recognition and object recognition\cite{hu2017deep}). Kumar et al.\cite{kumar2011co} developed a multi-view spectral clustering framework which achieves this goal by co-regularizing a clustering hypotheses, and proposed two co-regularization schemes to accomplish this.

As for high-dimensional data can be different to interpret, little progress has been made in dimension reduction across multiple views\cite{xia2010multiview}. We consider transforming data from the high-dimensional space to subspace with lower dimensions. The data transformation may be linear, as in principal component analysis (PCA)\cite{wold1987principal}, et al.\cite{feng2015locality}, but many nonlinear dimensionality reduction techniques also exist, which assume that the data of interest lie on an embedded non-linear manifold within the higher-dimensional space, such as Laplacian Eigenmaps (LE)\cite{belkin2003laplacian} and Locally Linear Embedding (LLE)\cite{roweis2000nonlinear}, etc.

In this paper, we propose a novel multi-view dimension reduction method named Multi-view Reconstructive Preserving Embedding (MRPE) which can deal with multi-view features from nonlinear space. MRPE fully exploits correlations between samples using linear reconstruction from multiple views. Meanwhile, MRPE constructs an optimization problem which can force multi-view features to learn from each other. Then, an iterative procedure is described in our paper to show the optimization process of MRPE. The contributions of our paper can be listed as follows: Firstly, we carefully survey the field of multi-view learning and propose a multi-view dimension reduction method. Secondly, MRPE fully exploits the local symmetries of linear reconstruction from multiple views, which is capable of learning the global structure of nonlinear manifold from various perspectives. Various experiments have verified that our proposed MRPE achieves excellent performance in most situations.

This paper is organized as follows: in section 2, we introduce some related works which consist of LLE and some basic knowledge of multi-view learning. And we describe the constructing procedure and the optimization procedure of our proposed MRPE in section 3. In section 4, we conduct several experiments on various multi-view datasets to verify the excellent performance of MRPE. And we make a conclusion in section 5.

\section{Related Works}
In this section, first, we provide a brief overview of LLE which is for single view data. Then, we introduce the background of multi-view learning in detail to help readers improve their ability to comprehend this paper.

\subsection{Locally Linear Embedding}
Locally linear Embedding (LLE) is a well-known nonlinear dimension reduction method which has attracted widely attentions. LLE begins by finding a set of the nearest neighbors for each sample, and then computes a set of weights for each sample as a linear combination of its neighbors, finally uses an eigenvector-based optimization technique to find the low-dimensional embedding of samples, such that each sample is still described with the linear combination of its neighbors. In order to find the low-dimensional embedding, LLE constructs the objective function as follows:

\begin{equation}\label{eq1}
\begin{aligned}
\underset{Y} {argmin} \quad & tr(YMY^T) \\
s.t. \quad & YY^T = I
\end{aligned}
\end{equation}
Where the matrix $M$ can be stored and manipulated as the sparse matrix $(I-W)(I-W)^T$ whose minimum $d$ nonzero eigenvectors provide the final low-dimensional representations $Y$. $W$ consists of linear correlations between each sample and its neighbors in the original space. Because LLE aims to maintain the linear correlations between samples,  $W$ is also the reflection of the similarities among each sample and its neighbors in lower space. Therefore, $W$ is a $n \times n$ sparse matrix which represents the same weight preserved within the original space and lower space. $tr(\cdot)$ represents the trace of the matrix. The dimensionality reduction by LLE succeeds in identifying the underlying structure of the manifold and is capable of generating highly nonlinear embeddings\cite{saul2000introduction}.

\subsection{Multi-view Learning}

Multi-view learning is concerned with the problem of machine learning from data represented by multiple distinct feature sets. This learning mechanism is largely motivated by the property of data from real-world where examples are described by different feature sets or different views in high-dimensional space\cite{sun2013survey}. For instance, as Fig.\ref{fig1}, one image can be represented by vectors extracted by multiple feature descriptors. For example, Gist\cite{oliva2001modeling} is a feature descriptor which can reflect the information of a region boundary of the object of shape of the scene in the images. Histogram of oriented gradients (HOG)\cite{dalal2005histograms} counts the occurrences of gradient orientation in localized portions. Local binary patterns (LBP)\cite{ojala2002multiresolution} is a simple yet efficient operator ot describe local image pattern and achieves good performances in most tasks. As we can see, features extracted from different descriptors show multiple properties which provide compatible and complementary information. However, features from multiple views always locate in different dimensions which can not be utilized directly. Therefore, by utilizing the consistency and complementary properties of different views, multi-view learning is rendered more effective, more promising, and has better generalization ability than single view learning.

For multi-view learning methods, we provide significant symbols utilized in the rest of our paper in order to present the technique details effectively. Capital letters, e.g., $X$, represent matrices or sets. Lower case letters, e.g., $x$, represent vectors, and $x_i$ is the $i$th item of $x$. Superscript, e.g., $X^{(v)}$, represents data from the $v$th view. Suppose multi-view features with $n$ samples having $m$ views can be represented as $X = \{X^{(v)} \in R^{D_v \times n}\}_{v=1}^m$, where $X^{(v)} = [x_1^{(v)}, x_2^{(v)}, \dots, x_n^{(v)}] \in R^{D_v \times n}$. And $X^{(v)}$ is the feature matrix which consists of features from the $v$th view. $D_v$ is the dimension of features in original space. The goal of our algorithm is to reduce the dimensionality to $d$ such that $D_v >> d$ and obtain the common low-dimensional representations as $Y = [y_1, y_2, \dots, y_n] \in R^{d \times n}$ which incorporates the feature information from multiple perspectives.

Multi-view learning has been a hot topic for a long time. There are many useful works \cite{Wang2015Robust,Wangyang2017Multi} which can deal with features from multiple views well. Co-regularized framework\cite{kumar2011co} is a good method which can deal with multi-view clustering by minimizing the distinguish between different views. The comparing methods(Pairwise and Centroid) in our experiment all adopt the co-regularized framework with different constraints. MSE\cite{xia2010multiview} is another useful work which can obtain one common low-dimensional representation for features from different views. Meanwhile, some co-training methods\cite{Kumar2011A} have also attracted attentions from researchers all over the world. Therefore, multi-view learning has been a hot topic during the last decades, which should be carefully studied.

\section{Method}

\begin{figure}[htbp]
\centering
\includegraphics[width=\textwidth]{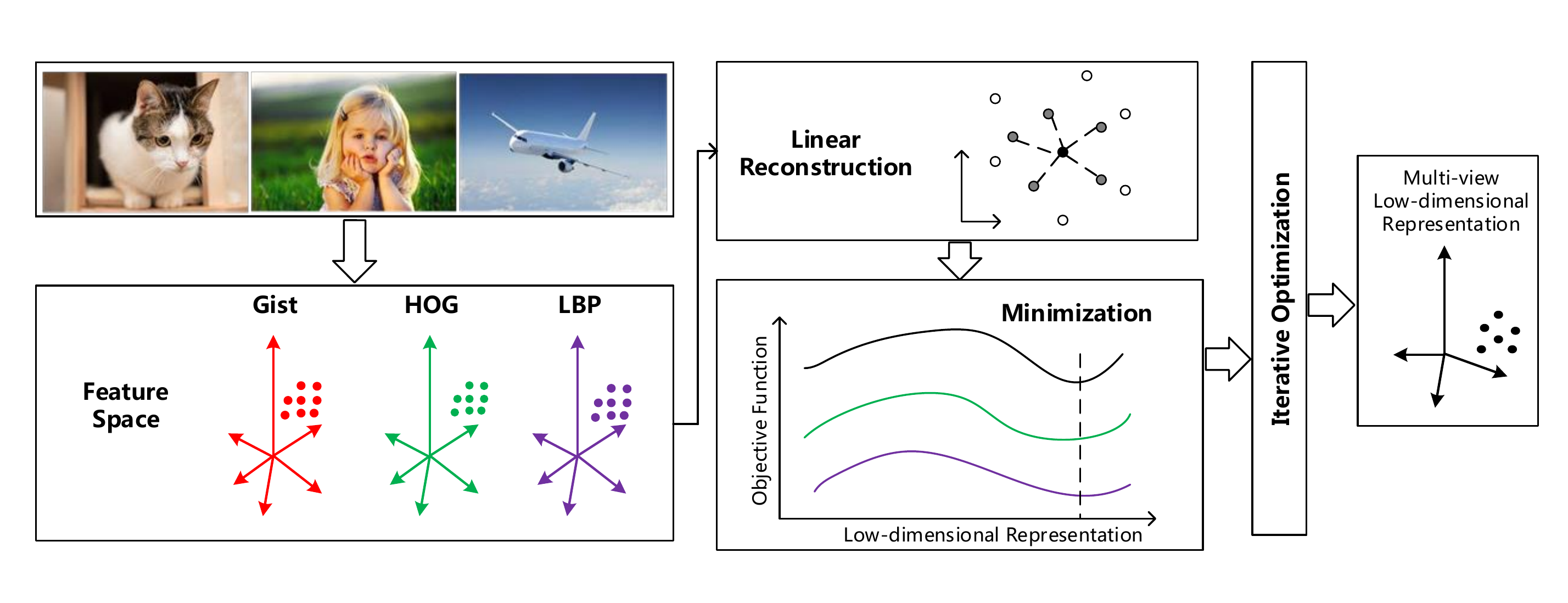}
\caption
{ \label{fig1}
The working procedure of MRPE. }
\end{figure}

In this section, we present our proposed MRPE which finds a common low-dimensional embedding over all views simultaneously, for we assume that different views hold the common and consistent low-dimensional representations. Fig.\ref{fig1} shows the working procedure of MRPE. First, MRPE exploits correlations among samples via linear reconstruction across multiple views. Meanwhile, MRPE constructs an optimizaiton problem which makes multi-view features learn from each other. Then, the iterative optimization is performed for MRPE to get the optimal and common low-dimensional representations from multiple views.

\subsection{Constructing Procedure}

In this paper, in addition to considering the distances between each two samples, we pay more attention to the similarities between each sample and its neighbors. Firstly, we reconstruct each sample by utilizing its $k$ neighbors to illustrate the similarities between features from each single view. Then, we map the similarities into lower-dimensional space so that it can preserve the underlying neighborhood structure of the original manifold. Finally, we develop a multi-view scheme to combine information from multiple views and achieve a better low-dimensional representation for samples.

First, we introduce how we obtain low-dimensional embedding for single view. As we know, one object can be represented by different views. We desire to integrate the features from multiple perspectives, so we expand the scheme of single view into multi-view learning and propose the approach MRPE in this paper for multi-view dimension reduction. Suppose the data consists of $n$ real-valued vectors $x_i^{(v)} \, (1 \leq i \leq n)$. And each $x_i^{(v)}$ represents the $i$th sample from the $v$th view with the dimensionality $D$. Therefore, the $i$th sample can be reconstructed from its $k$ nearest neighbors $x_{ij}^{(v)} \, (1 \leq j \leq k)$ using linear combination. So we minimize the cost function subject to measure the reconstruction errors of single view:

\begin{equation}\label{eq2}
\underset{w_{ij}^{(v)}}{argmin} \quad \sum_{i=1}^n ||x_i^{(v)} - \sum_{j=1}^k w_{ij}^{(v)} x_{ij}^{(v)}||^2
\end{equation}

Take the constraint $\sum_j^k w_{ij}^{(v)} = 1$ into consideration, we can compute the weight $w_{ij}^{(v)}$ to reconstruct the same sample in lower $d$ dimensional space. The solving procedure of $w_{ij}^{(v)}$ is same with that of LLE, which can be found in \cite{roweis2000nonlinear}.  $w_{ij}^{(v)}$ maintains the correlation between the $i$th and $j$th samples. It's essential to fully exploit correlations between samples via linear reconstruction and maintain them into the low-dimensional representations. So each sample $x_i^{(v)}$ in original space is mapped into a sample $y_i$ in lower dimensional space by minimizing the objective function:

\begin{equation} \label{eq3}
\begin{aligned}
\underset{y_i}{argmin} \quad & \sum_{i=1}^n ||y_i - \sum_{j=1}^k w_{ij}^{(v)} y_{ij}||^2 \\
s.t. \quad & \sum_i^n y_i = 0; \frac{1}{n} \sum_i^n y_i y_i^T = I
\end{aligned}
\end{equation}
where the optimal $y_i$ can be achieved with the weight $w_{ij}^{(v)}$ maintained in lower-dimensional space. $w_{ij}^{(v)}$ can be obtained by Eq.\ref{eq2}. Concerning the descriptions of one object in multiple views, we desire to integrate the features from various perspectives. Therefore, our approach constructs the following objective function according to Eq.\ref{eq3} by summing all the views which gives contribution to multi-view low-dimensional embedding as follows:

\begin{equation} \label{eq4}
\begin{aligned}
\underset{y_i}{argmin} \quad \sum_{v=1}^m \sum_{i=1}^n \alpha_v ||y_i - \sum_{j=1}^k w_{ij}^{(v)} y_{ij}||^2.
\end{aligned}
\end{equation}

As we mentioned above, multiple properties from multiple views are the descriptions of exactly one same object, and due to their complementary information of multiple views to each other, different views definitely have different contributions to the final common low-dimensional embedding. In order to fully explore complementary properties of different views, a set of nonnegative weights $\alpha = [\alpha_1, \dots, \alpha_m]$ is imposed on part optimizations of different views independently. Since the coefficients shows the different weights of each view, naturally the larger $\alpha_v$ is, the more crucial role the view $X^{(v)}$ plays in learning to procure the common low-dimensional embedding $Y$ across multiple views. Moreover, we add the constraint $\sum_{v=1}^m \alpha_v = 1$. In order to obtain a more compact expression, Eq.\ref{eq4} can be reorganized as follows and the inference process can be found in Appendix:

\begin{equation} \label{eq5}
\begin{aligned}
\underset{Y, \alpha}{argmin} \quad & \sum_{v=1}^m \alpha_v tr(YM^{(v)}Y^T) \\
s.t. \quad & YY^T = I; \sum_{v=1}^m\alpha_v = 1, \alpha_v \geq 0.
\end{aligned}
\end{equation}
where $M^{(v)}$ is a $n \times n$ matrix found as $M^{(v)} = (I-W^{(v)})(I-W^{(v)})^T$ whose minimum d nonzero eigenvectors provide the final low dimensional representations $Y = [y_1, \dots, y_n] \in R^{d \times n}$. Sparse matrix $W^{(v)} \in R^{n \times n}$ describes the reconstructing weights among the samples. The constraint $YY^T = I$ is imposed on Ref.~\cite{roweis2000nonlinear} to uniquely determine the common low-dimensional embedding $Y$ of multiple views.

By means of analyzing the equation Eq.\ref{eq5}, we notice that the solution to $\alpha$ in Eq.\ref{eq5} is $\alpha_t = 1$ corresponding to the minimum $tr(YM^{(v)}Y^T)$ over different views, and $\alpha_t = 0$ otherwise. This solution suggests that only exactly one view is finally selected by this method. Therefore, the performance of this method is equivalent to the one from the best view which definitely does not meet our goal to integrate compatible and complementary properties from multiple views to obtain better performance and get a better embedding than based on just one single view. So we adopt a trick utilized in Ref.\cite{wang2007optimizing} to avoid this circumstance. We set $\alpha_v$ to $\alpha_v^r$ with the constraint $r > 1$. In this condition, $\sum_{v=1}^m \alpha_v^r = 1$ carries out its minimum when $\alpha_v = 1/m$ with respect to $\sum_{v=1}^m \alpha_v = 1, \alpha_v > 0$. By setting $r > 1$, Similar $\alpha_v$ for different views will be achieved. Therefore, each single view has a particular contribution to the final low-dimensional embedding $Y$. Thus, the new objective function is defined as

\begin{equation} \label{eq6}
\begin{aligned}
\underset{Y, \alpha}{argmin} \quad & \sum_{v=1}^m \alpha_v^r tr(YM^{(v)}Y^T) \\
s.t. \quad & YY^T = I; \sum_{v=1}^m\alpha_v = 1, \alpha_v \geq 0.
\end{aligned}
\end{equation}

According to the related discussions, MRPE finds a low-dimensional sufficiently smooth embedding $Y$ by preserving the underlying neighborhood similarity of the original manifold in each view simultaneously. Furthermore, we illustrate the optimization procedure of MRPE in the next part.

\subsection{Optimization Procedure}

In this section, we introduce the iterative optimization procedure of MRPE in detail. It is clearly that MRPE aims to obtain the common low-dimensional representations from multiple views which fully utilize information from the other views. In this paper, we derive an iterative algorithm by utilizing the alternating optimization to achieve the optimal solution which iteratively updates $Y$ and $\alpha$.

Firstly, we fix $Y$ to update $\alpha$. By adopting a Lagrange multiplier $\lambda$ to take the constraint $\sum_{v=1}^m \alpha_v = 1$ into consideration, we get the Lagrange function

\begin{equation}
L(\alpha, \lambda) = \sum_{v=1}^m \alpha_v^r tr(YM^{(v)}Y^T) - \lambda(\sum_{v=1}^m \alpha_v - 1).
\end{equation}

By means of setting the derivative of $L(\alpha, \lambda)$ with respect to $\alpha_v$ and $\lambda$ to zero, we can obtain

\begin{equation} \label{eq7}
\alpha_v = \frac{(1 / tr(Y M^{(v)} Y^T)) ^ {1 / (r-1)}}{\sum_{v=1}^m (1 / tr(Y M^{(v)} Y^T)) ^ {1 / (r-1)}}.
\end{equation}

According to Eq.\ref{eq7}, we have the following understanding for $r$ in controlling $\alpha_v$. If $r$ tends to $\infty$, different $\alpha_v$ will be close to each other. If $r$ tends to $1$, only $\alpha_v = 1$ corresponding to the minimum $tr(YM^{(v)}Y^T)$ over different views, and $\alpha_v = 0$ otherwise. Therefore, the selection of $r$ should be based on the complementary properties of all views. Abundant complementary features prefer larger parameter $r$; otherwise, $r$ should be small.

Secondly, we fix $\alpha$ to update $Y$, thus, the optimization problem in Eq.\ref{eq6} is equivalent to

\begin{equation} \label{eq8}
\begin{aligned}
\underset{Y}{min} \quad & tr(YMY^T) \\
s.t \quad & YY^T = I
\end{aligned}
\end{equation}
where $M = \sum_{v = 1}^m \alpha_v^r M^{(v)}$. The optimal $Y$ is given as the eigenvectors associated with the smallest $d$ eigenvalues of the sparse matrix $M$.

As stated in the aforementioned descriptions, we can form an iterative optimization procedure, presented in Algorithm 1 to obtain the optimal solution of MRPE.

\vspace{3ex}

\begin{tabularx}{0.9\textwidth}{ X }
\hrule
\textbf{Algorithm 1}: MRPE Algorithm \\
\hline
\textbf{Input}: \\
$X = \{X^{(v)} \in R^{D_v \times n}\}_{v=1}^m$: features from all views; $d$: the dimension of the low-dimensional embedding ($d < D_v, 1 \leq v \leq m$); $r > 1$. \\
\textbf{Output}: \\
$Y = [y_1, y_2, \dots, y_n] \in R^{d \times n}$: the low-dimensional representations for all views. \\
\hline
\textbf{Optimization Procedure}:
\begin{enumerate}
\item Calculate reconstructing weight matrix $W^{(v)}$ for each view.
\item Construct similarity matrix between each two samples.
\item Initialize parameters d, k, r and $\alpha = [1/m, \dots, 1/m]$.
\item Repeat.
\item Obtain the eigenvectors associated with the smallest $d$ eigenvalue of the matrix $M$ defined in Eq.\ref{eq6} and get the low-dimensional embedding $Y$.
\item Calculate and update $\alpha$.
\item Until convergence.
\end{enumerate}
\hrule
\end{tabularx}

\vspace{3ex}

The objective function $\sum_{v=1}^m \alpha_v^r tr(YM^{(v)}Y^T)$ reduces with the increasing of the iteration number so that our algorithm converges. In particular, with fixed $\alpha$, the optimal $Y$ can reduce the value of the objective function, and with fixed $Y$, the optimal $\alpha$ will also reduce the value of objective function.

\section{Experiment}

In this section, we present experiments on publicly available datasets for document classification, face recognition and image retrieval, which serve both to demonstrate the efficacy of the proposed multi-view dimension reduction method MRPE and validate the claims of the previous sections. We will first introduce the datasets and illustrate the comparing methods to evaluate our proposed method MRPE. We will then demonstrate various experiments, comparing performance across different feature dimensions in lower space, and show the accuracies on each dataset. All these experiments can verify that our proposed MRPE achieves better performance in most situations.

\subsection{Datasets and comparing methods}

\begin{figure*}[ht]
\centering
\subfloat[Yale Dataset]{\includegraphics[width=0.4\textwidth]{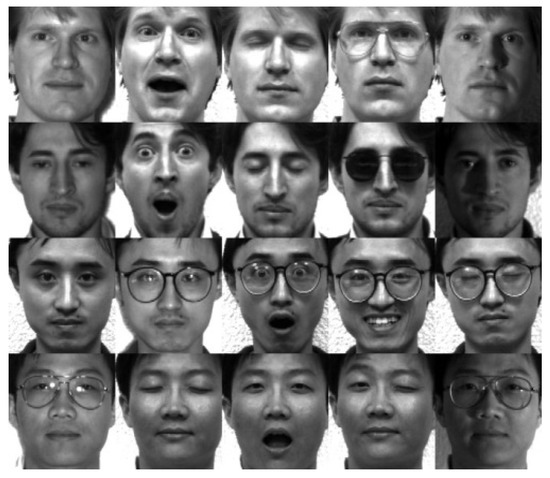}}
\subfloat[Corel-1K]{\includegraphics[width=0.6\textwidth]{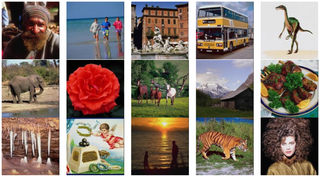}}

\subfloat[Holiday Dataset]{\includegraphics[width=\textwidth]{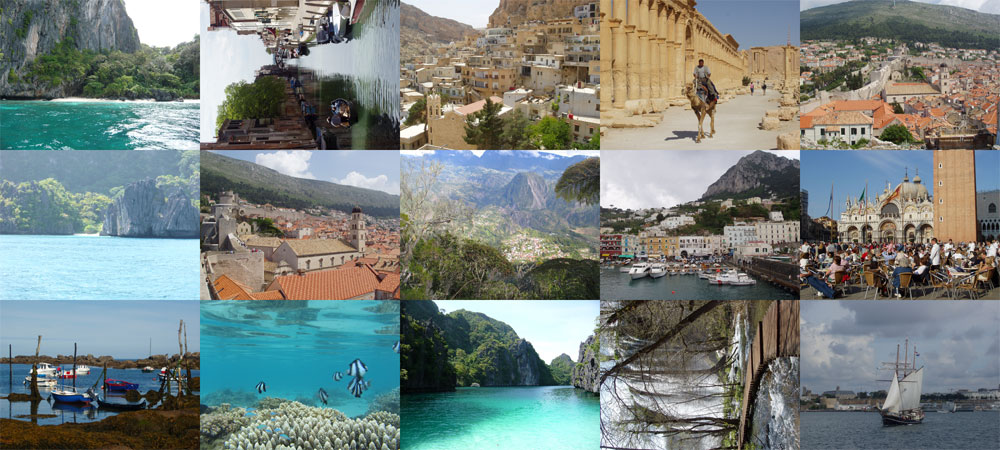}}

\caption
{ \label{fig2}
Images from datasets utilized in our experiments. }

\end{figure*}

In our experiments, 5 datasets are utilized to illustrate the effectiveness of MRPE, including document datasets (3sources \footnote{http://mlg.ucd.ie/datasets/3sources.html}) and face datasets (Yale \footnote{http://cvc.yale.edu/projects/yalefaces/yalefaces.html} and ORL \footnote{http://www.uk.research.att.com/facedatabase.html}), and image datasets (such as, Holidays \footnote{http://lear.inrialpes.fr/jegou/data.php} and Corel-1K \footnote{https://sites.google.com/site/dctresearch/Home/content-based-image-retrieval}). All document datasets are benchmark multi-view datasets. For those images datasets, we extract features using multiple descriptors as multi-view features for our experiments. Some images from these datasets are shown as Fig.\ref{fig2}.

The performance of MRPE is evaluated by comparing the following methods: 1. Pairwise \cite{kumar2011co}, which is a multi-view spectral embedding method using pairwise constraints. 2. Centroid \cite{kumar2011co}, which utilizes centroid constraints for multi-view spectral embedding. 3. MSE \cite{xia2010multiview}, which is a multi-view dimension reduction method utilizing global coordinate alignment. 4. SLE which is laplacian eigenmaps \cite{belkin2003laplacian} with the best single view. 5. SLLE is locally linear embedding \cite{roweis2000nonlinear} with the best single view. 6. FCLE connects all the features among the multiple views and use laplacian eigenmaps for single view embedding. 7. FCLLE is locally linear embedding with feature connection.

After using the dimension reduction methods above, we can obtain the low-dimensional representations for all views. We calculated all the experiment results on the low-dimensional representations from each single view. And the experiment results are the best ones from all views 

\subsection{Document classification}

\begin{figure*}[htb]

\centerline{
\subfloat[dimension=10]{\includegraphics[width=0.5\textwidth]{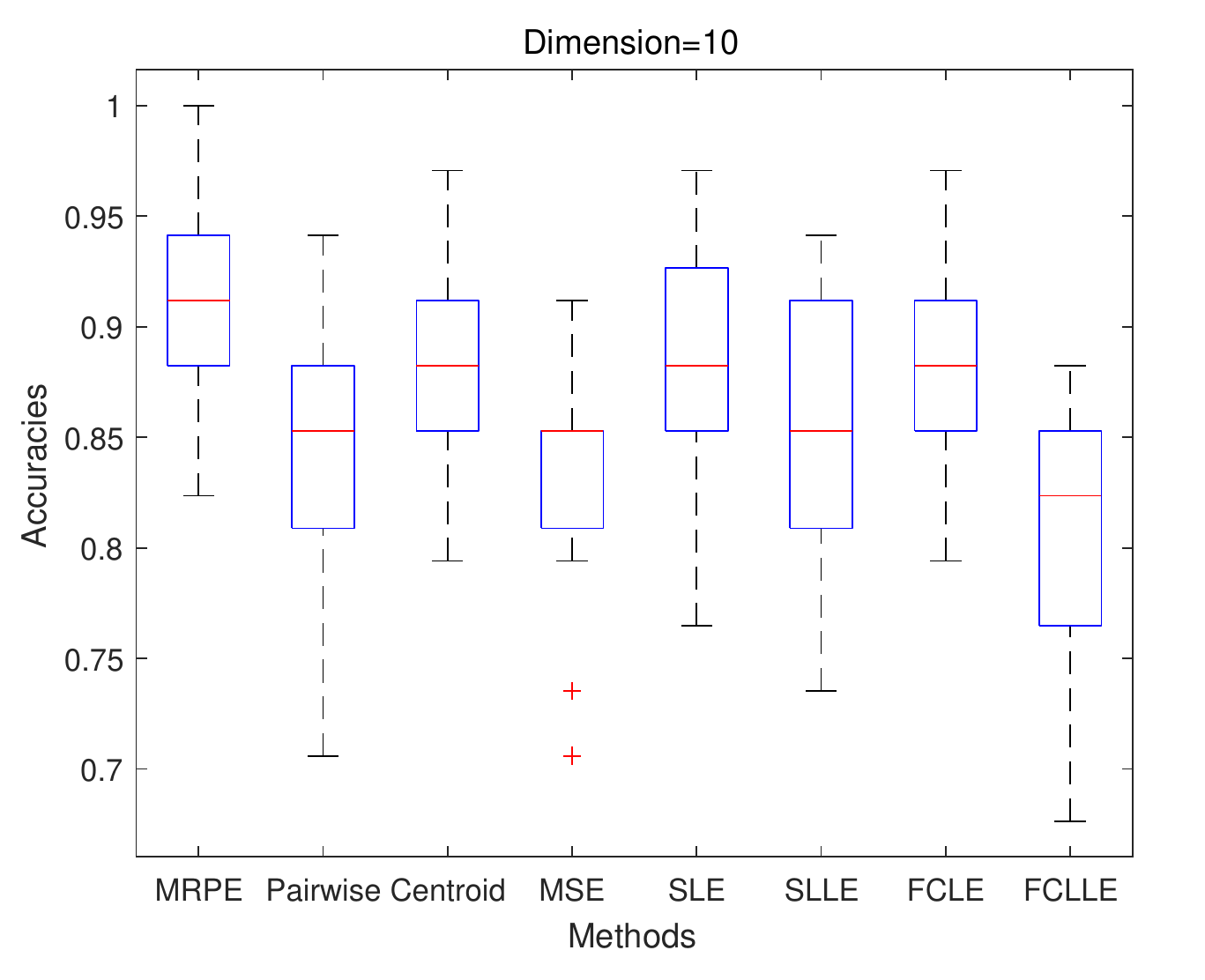}}
\subfloat[dimenion=20]{\includegraphics[width=0.5\textwidth]{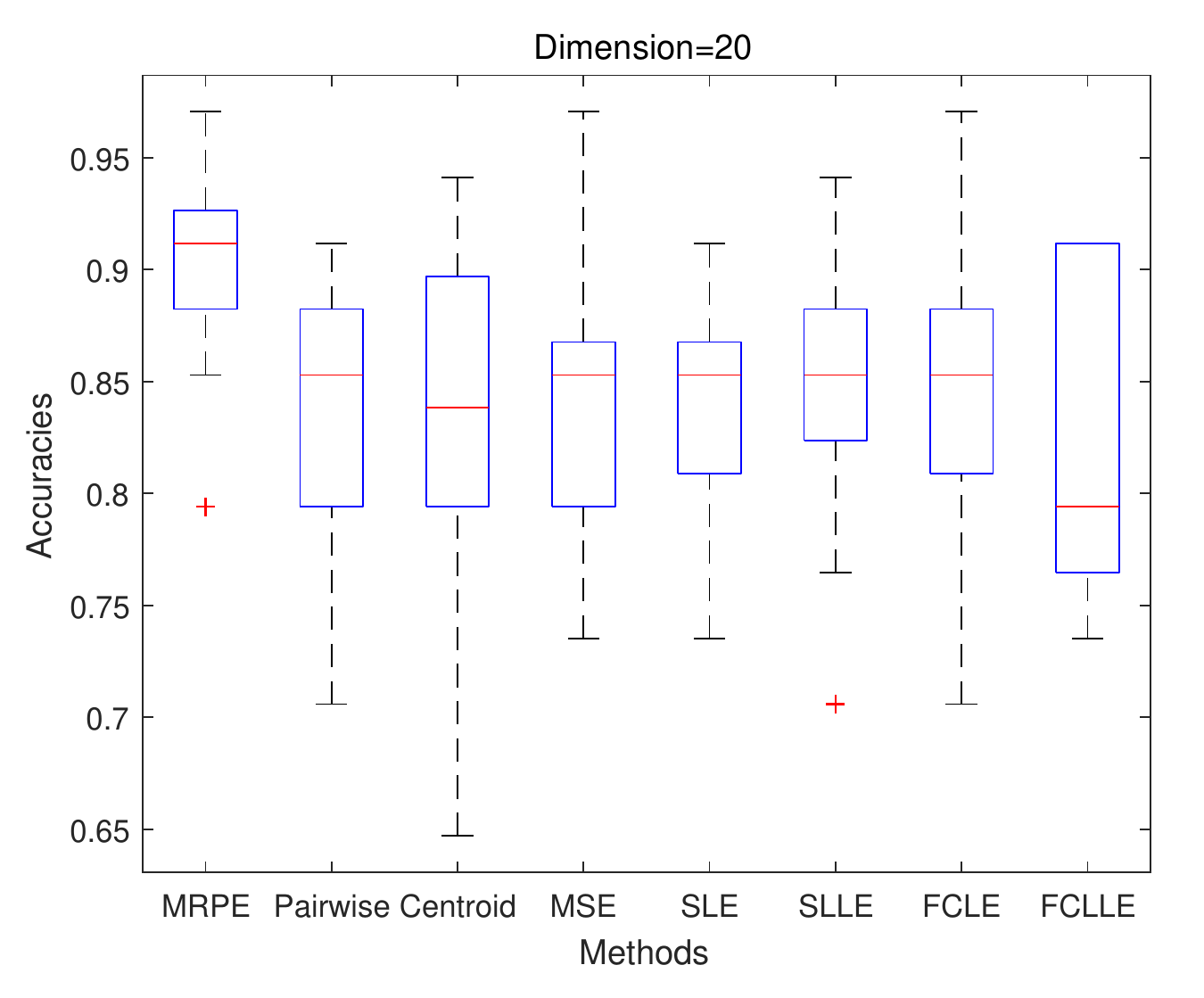}}
}
\caption{Classification Accuracies on 3Sources}
\label{fig3}
\end{figure*}

In this part, we carry out a document classification experiment on 3Sources dataset to show the effectiveness of our method. 3Sources is collected from three online new sources, BBC, Reuters and Guardian. All of the datasets include three views and each source is treated as one single view in this dataset. The demensions of features from these 3 views are 3068, 3631, 3560 respectively. Then, twenty percent of all samples are randomly assigned as those ones which need to be classified. MRPE and all comparing methods are trained to obtain low-dimension representations using the training samples. 1NN classifier \cite{altman1992introduction} is adopted to show the classification results on the testing samples. We conducted the experiment for twenty times and calculated the classification accuracies as the boxplot in Fig.\ref{fig3}.

As we can see, our proposed method MRPE achieve excellent performance to deal with the document dataset. Because 3Sources dataset consists of features from 3 views, MRPE can fully utilize all information from multiple views. Meanwhile, MREP maintains the linear reconstructive correlations between each sample and its neighbours, which improves the performances of it. Furthermore, we find out that the performances of those multi-view methods are much better than those singe-view ones, which can verify the value and potential of related researches on the field of multi-view learning. FCLE and FCLLE cannot achieve good performance, which verifies the infeasibility of feature concatenation.  

\subsection{Face Recognition}
\label{facerecognition}

\begin{figure*}[htb]
\centerline{
\subfloat[dimension=10]{\includegraphics[width=0.5\textwidth]{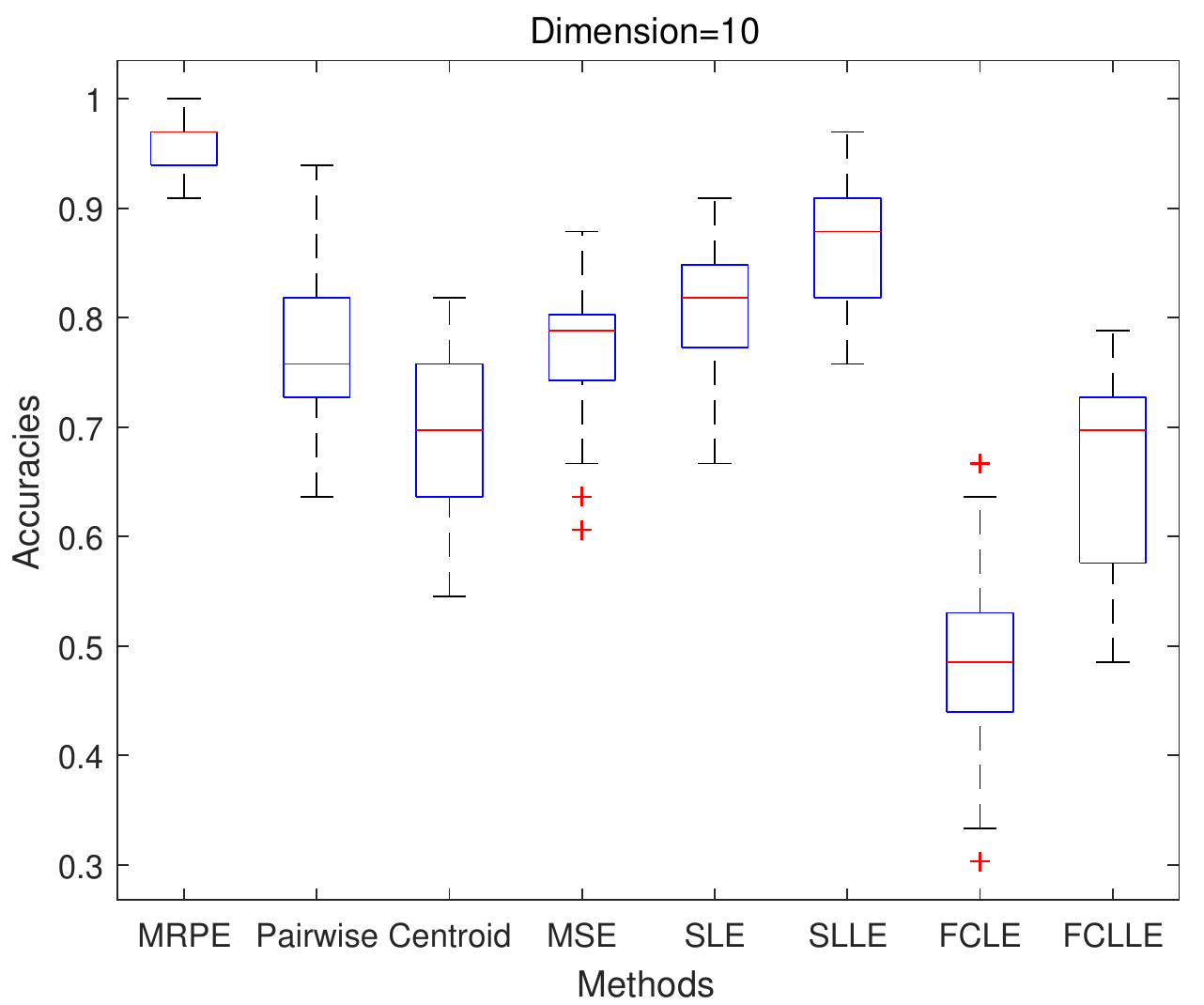}}
\subfloat[dimenion=20]{\includegraphics[width=0.5\textwidth]{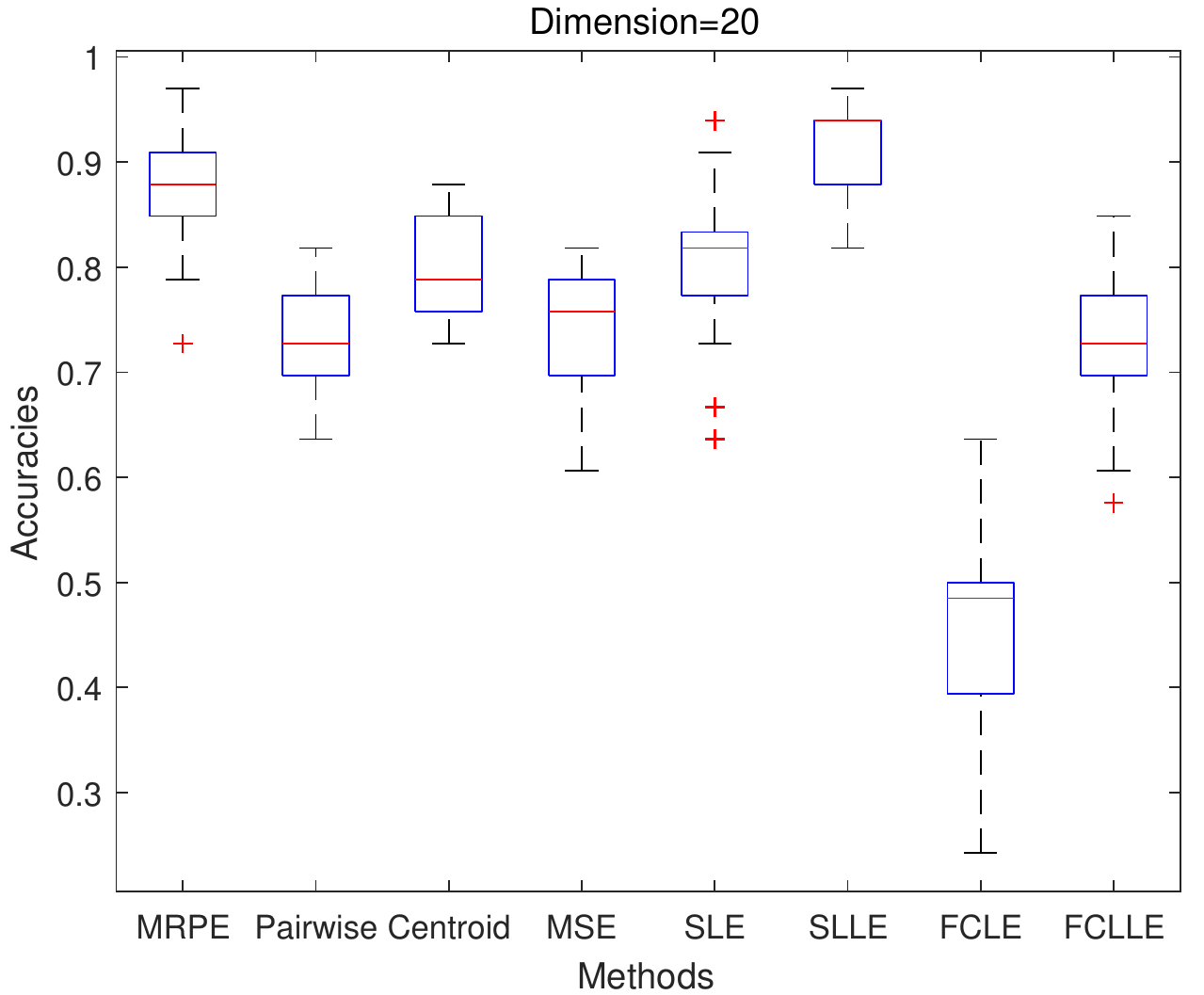}}
}
\caption
{ \label{fig4}
Recognition Accuracies on Yale Faces. }

\end{figure*}

In this part, we conduct face recognition experiments on Yale and ORL dataset to show the advantages of our proposed method MRPE. There are 165 faces corresponding to 15 people in Yale dataset. And there are 400 faces correspongding to 40 people in ORL dataset. We extract features by grey-scale intensity, local binary patterns \cite{ojala2002multiresolution} and edge direction histogram \cite{gao2008image} as three views. The demensions of features from these 3 views are 1024, 256, 72 respectively. Twenty percent faces are randomly selected to be recognized for both Yale and ORL datasets. 1NN classifier is adopted to calculate the recognition results after the dimension reduction. We conduct these two experiments for twenty times. For the experiment on Yale, we show the boxplots as Fig.\ref{fig4}. For ORL datasets, we summarized the recognition accuracies and show the mean and max results as Table.\ref{table1}.

\begin{table}[htbp]
	\centering
	\small
	\caption{Recognition accuracies on ORL dataset}
	\begin{tabular}
		{ccccccccc}
		\hline
		\textbf{ORL}& &	MRPE &	Pairwise &	Centroid &	MSE & SLE &	FCLE &	FCLLE  \\
		\hline
		\raisebox{-1.50ex}[0cm][0cm]{Dim=10 }&	Mean &	\textbf{83.71}{\%}&	82.56{\%}&	83.16{\%}&	82.97{\%}&	79.32{\%}&	80.79{\%}&	81.43{\%}  \\
		
		&	Max &	\textbf{87.62}{\%}&	85.48{\%}&	86.81{\%}&	85.79{\%}&	83.51{\%}&
		84.01{\%}&	84.75{\%}  \\
		\hline
		\raisebox{-1.50ex}[0cm][0cm]{Dim=20}&	Mean &	\textbf{85.58}{\%}&	84.33{\%}&	84.46{\%}&	84.06{\%}&	80.00{\%}&	81.32{\%}&	83.19{\%}  \\
		
		&	Max &	\textbf{88.26}{\%}&	86.06{\%}&	87.41{\%}&	87.30{\%}&	85.36{\%}&
		85.76{\%}&	86.20{\%}  \\
		\hline
	\end{tabular}
	\label{table1}
\end{table}

Fig.\ref{fig4} and Table.\ref{table1} can show that MRPE can achieve good performance in the field of face recognition. The performances of FCLE and FCLLE are bad because of their unreasonable way to deal with multi-view features. Furthermore, SLLE is another good method on Yale dataset. 

\subsection{Image Retrieval}

In this part, we conduct two experiments on different image datasets (including Holidays and Corel-1K datasets) for image retrieval, which brings excellent performance of MRPE.

For Holidays dataset, there are 1491 images corresponding to 500 categories, which are mainly captured for sceneries. Among all these images, 500 images are assigned as the query images while the other 991 are assigned as the corresponding relevant images. We employ MSD \cite{liu2011image}, Gist \cite{oliva2001modeling}, and HOC \cite{yu2016novel} to extract features as three views for all images. The demensions of features from these 3 views are 72, 512, 768 respectively. All these methods are conducted to project all samples into a 50 low-dimensional subspace. And distance metric \cite{wang2016semantic,wang2018learning} is essential for image retrieval and utilize $\ell^1$ distance to measure similarities between samples. We conduct this experiment for twenty times and show the precision, recall, mean average precision (MAP) and $F_1$-Measure on top 2 retrieval result as Table.\ref{table2}.

\begin{table}[htbp]
\caption{The precision (P\%), recall(R\%), MAP(\%) and $F_1-$Measure of different methods on Holidays dataset}
\begin{center}
\begin{tabular}{ccccccccc} \\
    \hline
    Criteria & MRPE & Pairwise & Centriod & MSE &
    SLE & SLLE & FCLE & FCLLE \\
    \hline
    P & \textbf{82.00} & 79.43 & 79.96 & 77.35 & 71.33 & 74.75 & 74.14 & 81.33 \\
    R & \textbf{63.42} & 61.23 & 61.59 & 59.76 & 54.68 & 57.39 & 56.98 & 62.89 \\
    MAP & \textbf{91.00} & 89.62 & 89.88 & 88.68 & 85.57 & 87.37 & 86.97 & 90.67 \\
    $F_1$ & \textbf{35.76} & 34.58 & 34.79 & 33.72 & 30.95 & 32.47 & 32.22 & 35.47 \\
    \hline
\end{tabular}
\end{center}
\label{table2}
\end{table}

\begin{figure*}[htb]
\centerline{
\subfloat[Precision]{\includegraphics[width=0.5\textwidth]{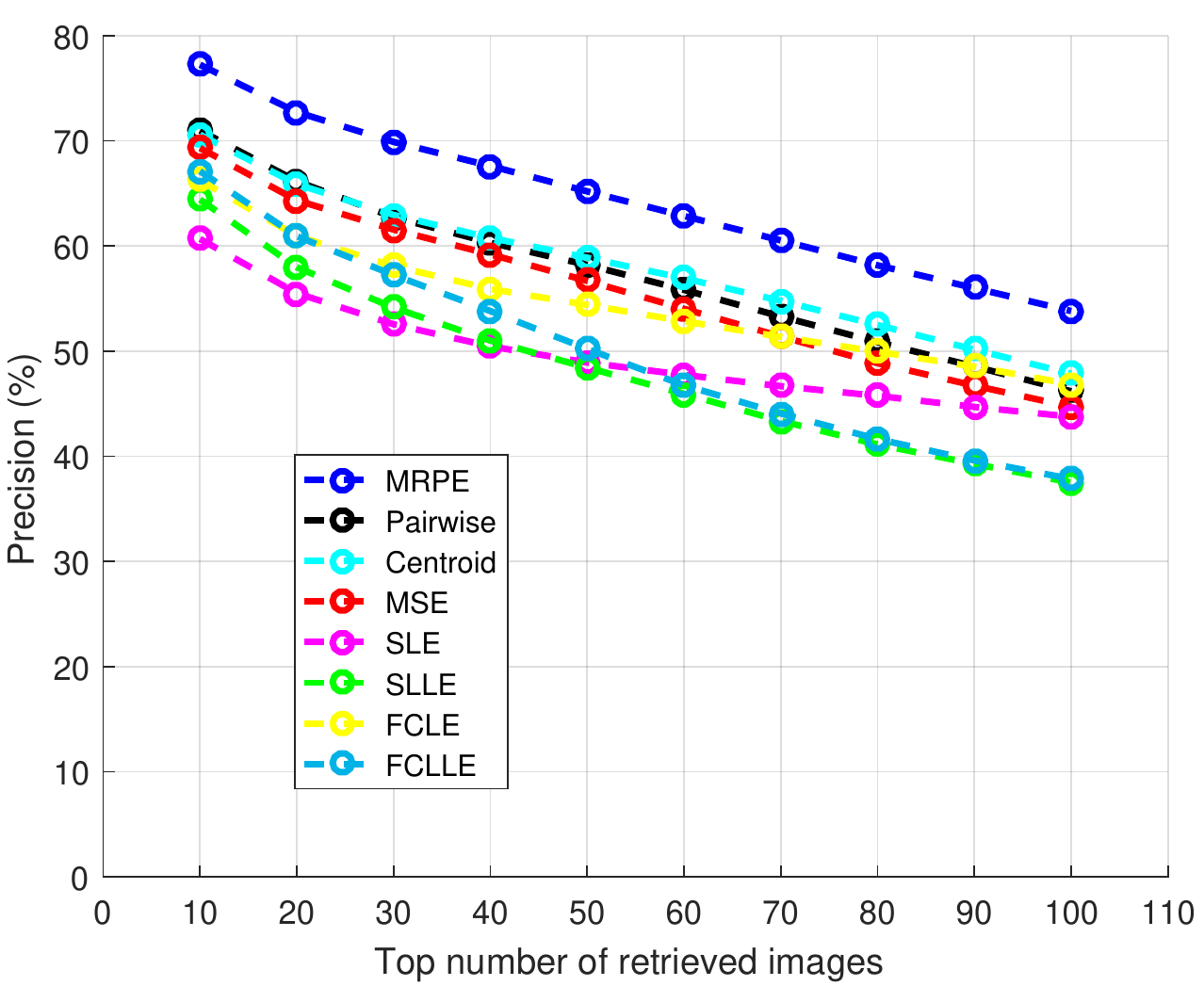}}
\subfloat[Recall]{\includegraphics[width=0.5\textwidth]{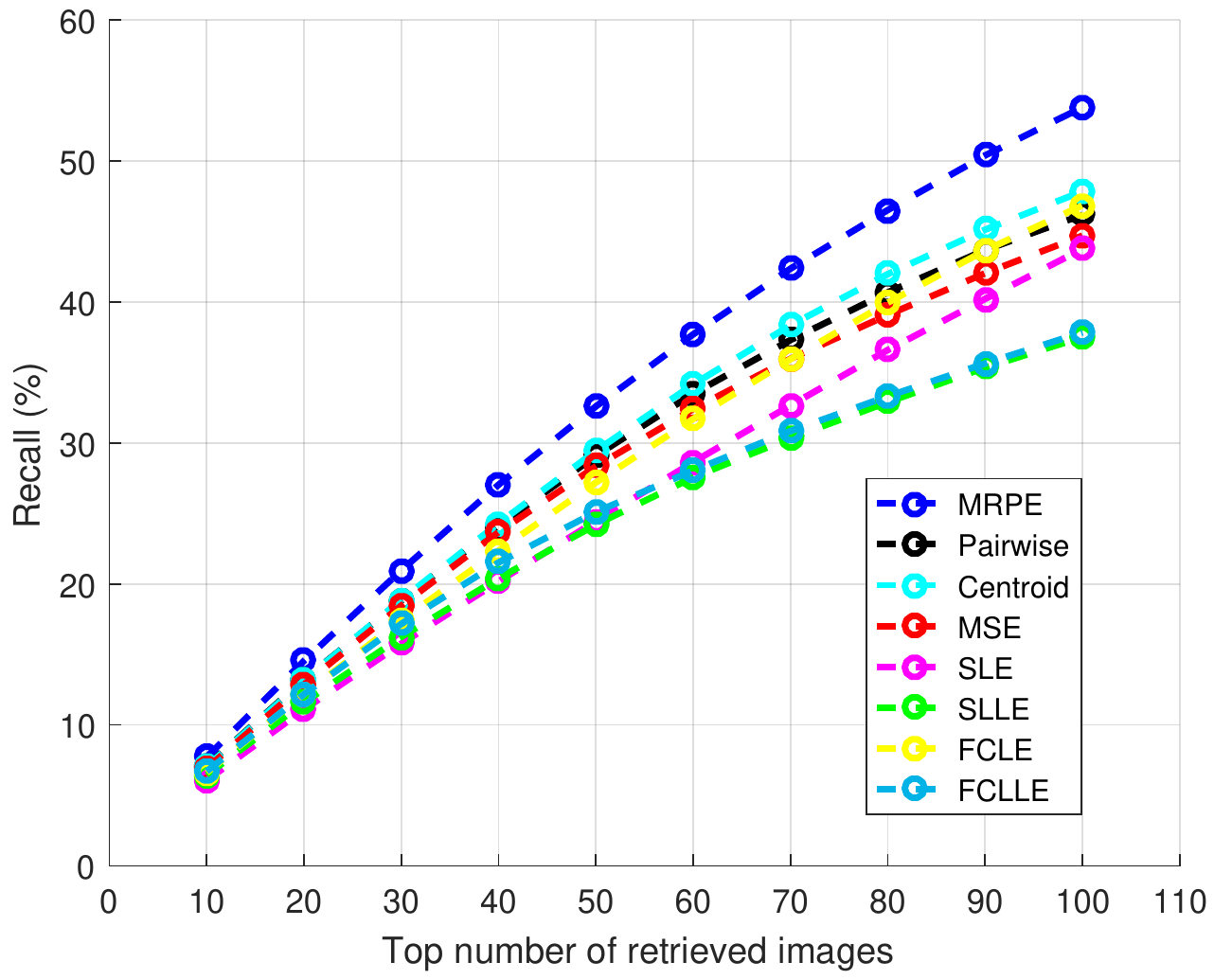}}}
\centerline{
\subfloat[PR-Curve]{\includegraphics[width=0.5\textwidth]{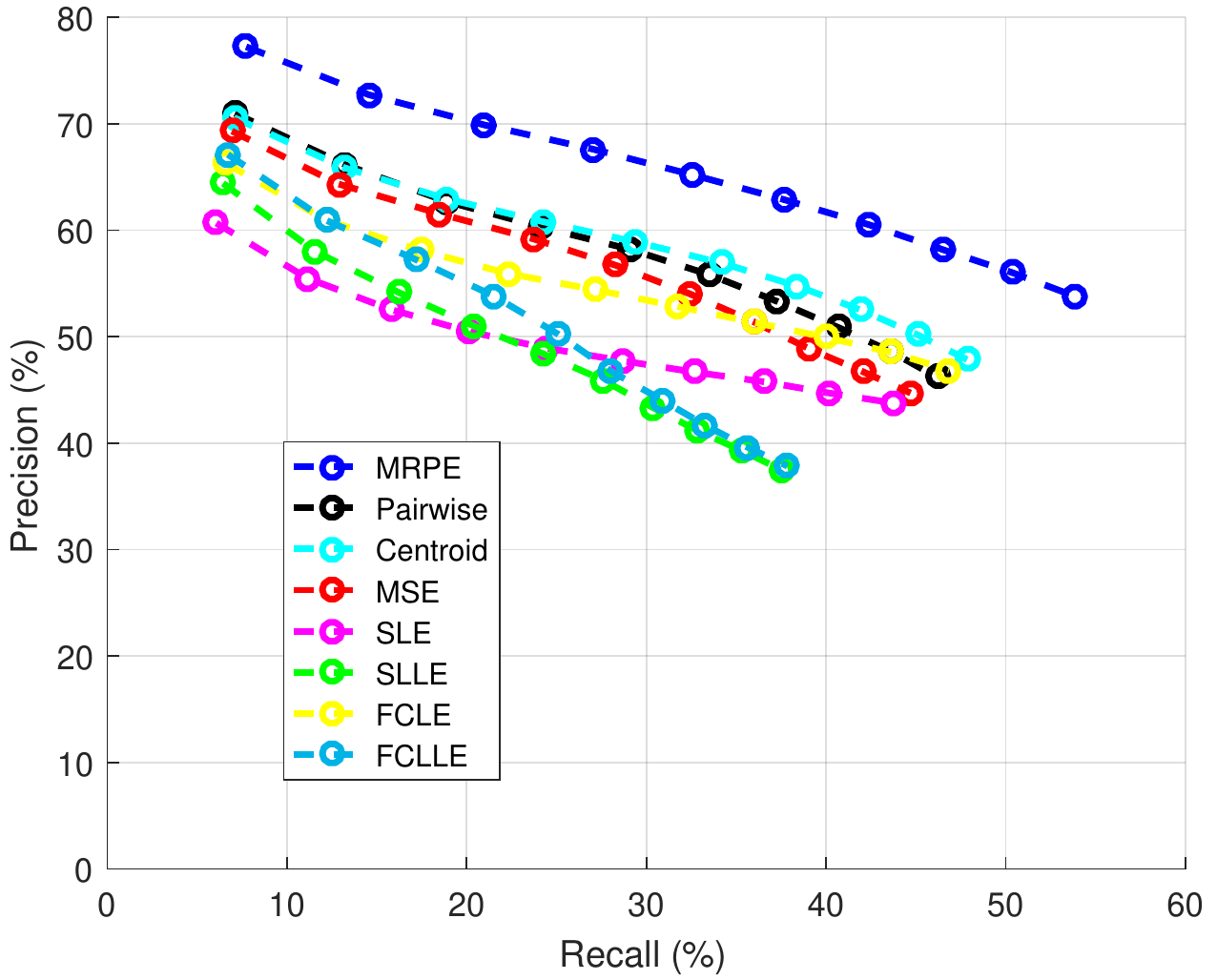}}
\subfloat[$F_1$-Measure]{\includegraphics[width=0.5\textwidth]{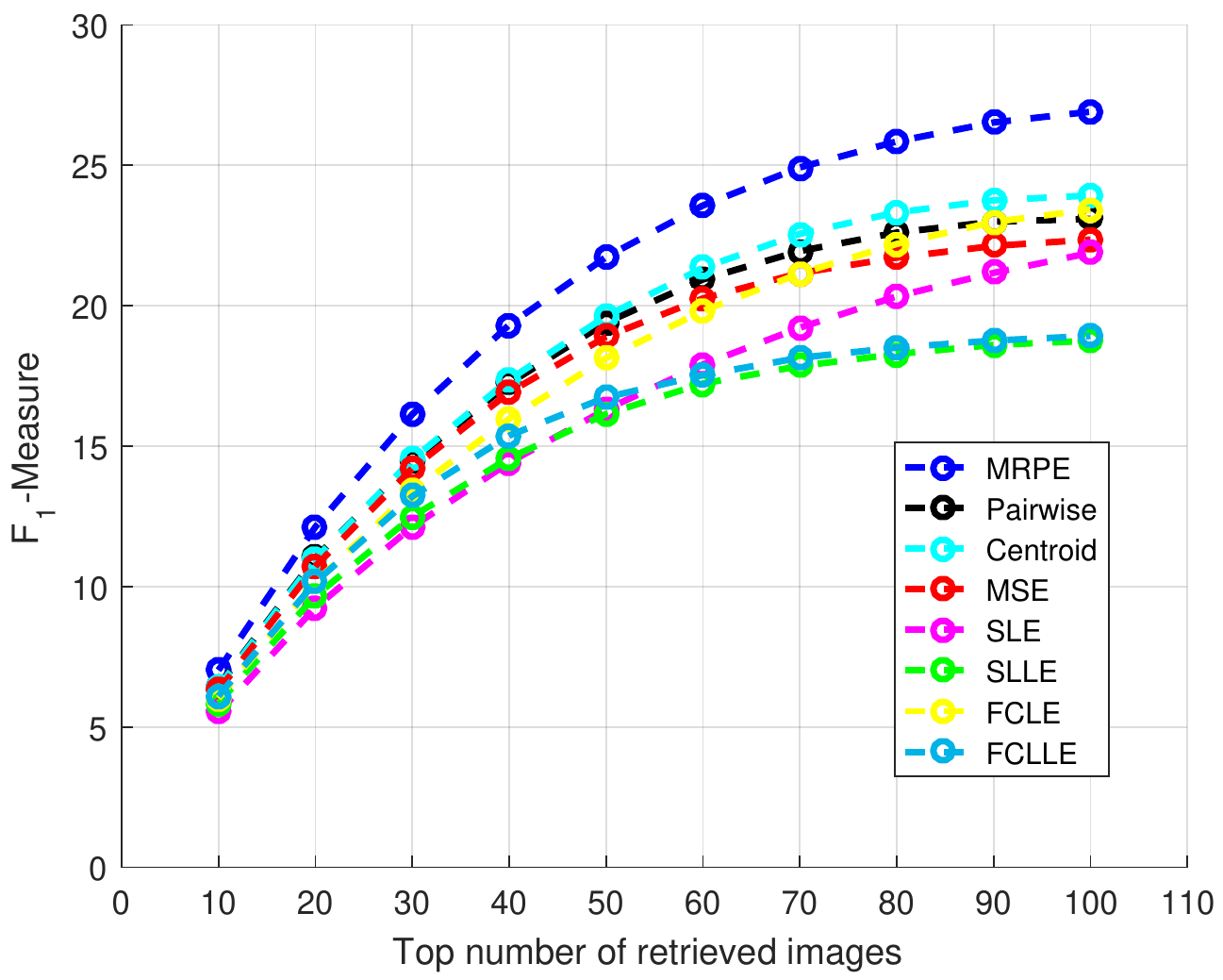}}
}
\caption{The curves of precision, recall, PR, and $F_1$-Measure on Corel-1k dataset}
\label{fig5}
\end{figure*}

For Corel-1K dataset, there are one thousand images corresponding to ten categories, which are collected just for image retrieval. There are 100 images for all categories. And for each category, we randomly  select 10 images as query ones. And we utilize the three descriptors \cite{liu2011image,oliva2001modeling,yu2016novel} to extract features same as above. The demensions of features from these 3 views are 72, 512, 768 respectively. Furthermore, all the methods are conducted to represent all samples in a 10 low-dimensional subspace. The procedure of this experiment is as same as that one on Holidays dataset. We conducted the experiment for twenty times and draw the relation curves as Fig.\ref{fig5}

Through these experiments for image retrieval, it can be found that our proposed Methods MRPE achieves the better performance than the other dimension reduction methods. Our proposed method MRPE can integrate compatible and complementary information from multiple views and obtain a better low-dimensional embedding from these views. In most situations, the first four multi-view methods outperform the other single view methods, which demonstrate that multi-view learning is an essential and valuable research field indeed.

\subsection{Convergence analysis and Training Time}

Because MRPE is solved by an iterative procedure, it is essential to discuss the convergence and training time of MRPE. In this section, we summarized the objective values and training time of the experiment on Yale dataset above. The dimension of the subspace is selected as 10. The training time was tested on a PC with a dual-Core i5-2300 CPU(2.80GHz) and 10 GB memory. All the details of this experiment has been shown in section \ref{facerecognition}. Fig.\ref{fig6} summarizes the the objective values and training time of on Yale dataset as follows:

\begin{figure}[htbp]
	\centering
	\includegraphics[width=0.6\textwidth]{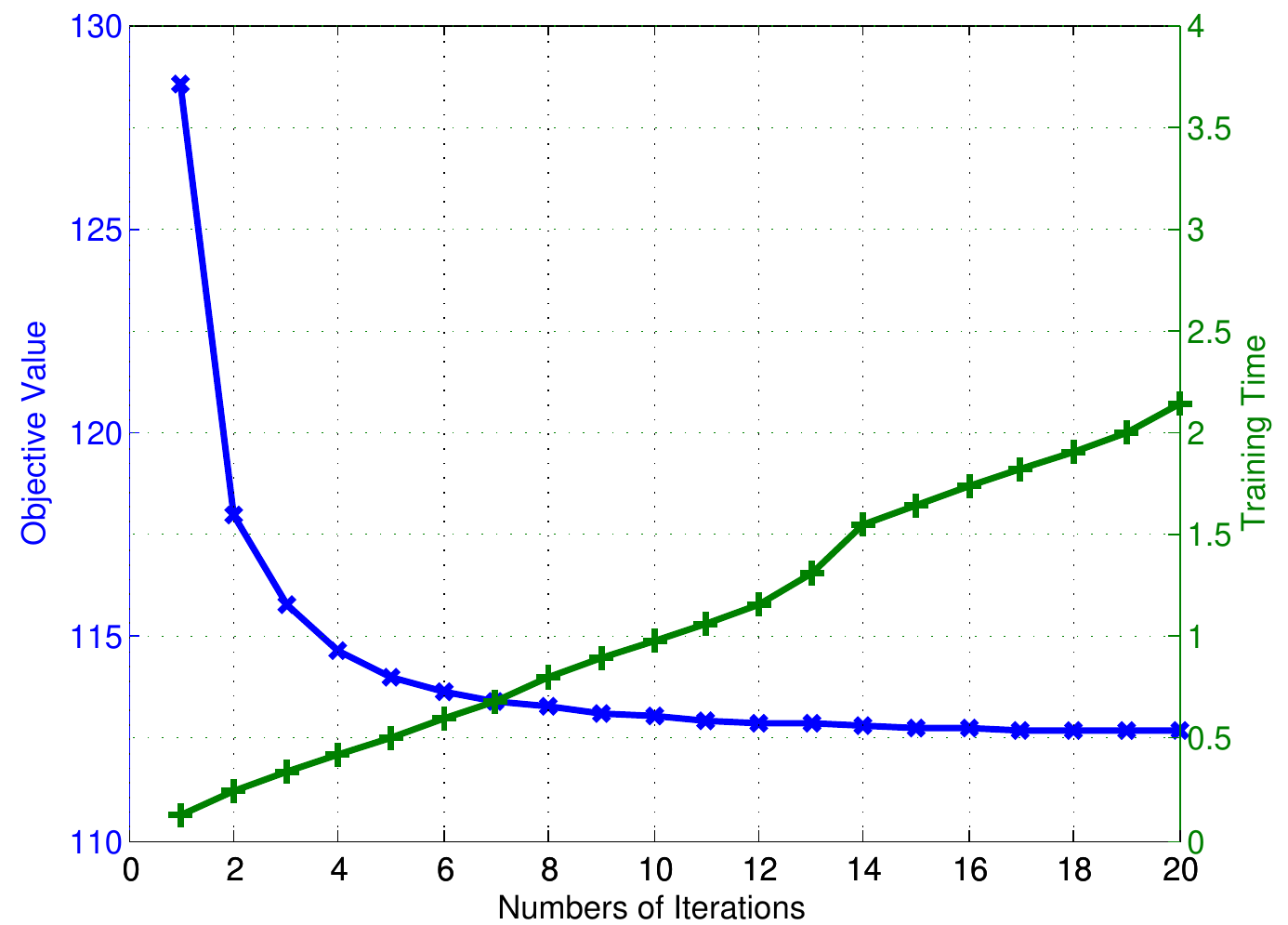}
	\caption
	{ \label{fig6}
		 Objective values and training time (in seconds) on Yale dataset.}
\end{figure}

It can be found that the objective values tend to be stable when 10 iterations are finished.  Therefore, the tendency of objective values verify that MRPE converges after enough iterations. Meanwhile, the consuming time of different iterations are almost the same with each other. Therefore, the training time of MRPE has linear correlation with the number of iterations.

\section{Conclusion}

In this paper, we propose a novel and effective multi-view dimension reduction method MRPE. MRPE reconstructs each sample by utilizing its $k$ neighbors to illustrate the similarities between each sample and its neighbors. The similarities is primely mapped into lower-dimensional space in order to preserve the underlying neighborhood structure of the original manifold. Therefore, MRPE can deal with different features from multiple views and fully exploit correlations between samples via linear reconstruction. Meanwhile, MRPE constructs an optimization problem which effectively explore and exploit multiple features simultaneously. And we evaluate the proposed approach on several real-world datasets via document classification, face recognition and image retrieval experiments and achieve excellent performance. 

Even though MRPE can achieve good performances in most stituation, there are still some sapces to improve its performance. Appropriate similarities between samples from multiple views can improve the discriminative ability of low-dimensional representations. Meanwhile, a good strategy to combine information from multiple views can improve the performances of most multi-view learning methods. Furthermore, the proposed MRPE cannot deal with mixed and incomplete data, which affects the  generalization of it. Therefore, our future works aim to improve the performances of MRPE and extend it to adapt various data (mixed or incomplete data).

\section{Compliance with Ethical Standards}

This study was funded by National Natural Science Foundation of P.R. China
(61173163, 61370200). Huibing Wang, Lin Feng, Adong Kong and Bo Jin declare that
they have no conflict of interest. This article does not contain any studies
with human participants or animals performed by any of the authors.

\section*{Appendix}

This appendix shows how to obtain Eq.\ref{eq5} from Eq.\ref{eq4} by matrix operations.

\begin{equation}
\begin{aligned}
& \sum_{v=1}^m \sum_{i=1}^n \alpha^{(v)} ||y_i - \sum_{j=1}^k w_{ij}^{(v)} y_{ij}||^2 \\
= & \sum_{v=1}^m \sum_{i=1}^n \alpha^{(v)} ||Y I_i - Y W_i^{(v)}||^2 \\
\end{aligned}
\end{equation}
where $Y = [y_1, \dots, y_i, \dots, y_n] \in R^{d \times n} $ is the final low-dimensional representations, $I_i = [0, \dots, 1, \dots, 0] \in R^{n \times 1}$ is a vector with value $1$ of the $i$th position, and $W_i^{(v)} \in R^{n \times 1}$ is extended by filling number $0$ of the vector $\sum_j^k w_{ij}^{(v)} \in R^{k \times 1}$.

\begin{equation}
\begin{aligned}
& \sum_{v=1}^m \sum_{i=1}^n \alpha^{(v)} ||y_i - \sum_{j=1}^k w_{ij}^{(v)} y_{ij}||^2 \\
= & \sum_{v=1}^m \sum_{i=1}^n \alpha^{(v)} ||Y I_i - Y W_i^{(v)}||^2 \\
= & \sum_{v=1}^m \sum_{i=1}^n \alpha^{(v)} (Y I_i - Y W_i^{(v)})(Y I_i - Y W_i^{(v)})^T \\
= & \sum_{v=1}^m \alpha^{(v)} tr((Y I - YW^{(v)})(Y I - YW^{(v)})^T) \\
= & \sum_{v=1}^m \alpha^{(v)} tr(Y(I-W^{(v)})(I-W^{(v)})^T Y^T) \\
= & \sum_{v=1}^m \alpha^{(v)} tr(YM^{(v)}Y^T)
\end{aligned}
\end{equation}
where $M^{(v)} = (I-W^{(v)})(I-W^{(v)})^T \in R^{n \times n}$ and $W^{(v)} = [W_1^{(v)}, \dots, W_n^{(v)}] \in R^{n \times n}$.


\bibliographystyle{unsrt}
\bibliography{report}   


\end{document}